# Bounded Conditioning:
# Flexible Inference for Decisions Under Scarce Resources*

**Eric J. Horvitz**    **H. Jacques Suermondt**
**Gregory F. Cooper**
Medical Computer Science Group
Knowledge Systems Laboratory
Departments of Computer Science and Medicine
Stanford, California 94305

## Abstract

We introduce a graceful approach to probabilistic inference called *bounded conditioning*. Bounded conditioning monotonically refines the bounds on posterior probabilities in a belief network with computation, and converges on final probabilities of interest with the allocation of a complete resource fraction. The approach allows a reasoner to exchange arbitrary quantities of computational resource for incremental gains in inference quality. As such, bounded conditioning holds promise as a useful inference technique for reasoning under the general conditions of uncertain and varying reasoning resources. The algorithm solves a probabilistic bounding problem in complex belief networks by breaking the problem into a set of mutually exclusive, tractable subproblems and ordering their solution by the expected effect that each subproblem will have on the final answer. We introduce the algorithm, discuss its characterization, and present its performance on several belief networks, including a complex model for reasoning about problems in intensive-care medicine.

## 1   Toward Flexible Inference and Representations

We have pursued the construction of representations and inference strategies that degrade gracefully, and in a well-characterized manner, as the amount of computation applied to reasoning is decreased. In general, the difficulty of a decision problem, the value of solving the problem, and the costs and availability of computation time may vary greatly. We desire inference strategies that are relatively insensitive to small changes in the amount of committed computation time. In particular, strategies that demonstrate (1) continuity and (2) monotonicity along a dimension of refinement, and that (3) converge on an optimal result in the limit of sufficient resources, are extremely valuable [7]. Such incremental-refinement approaches are of crucial importance for intelligent agents that depend on computation for decision making under the general condition of varying and uncertain resources because they allow for value to be extracted from incomplete solutions [7,5]. The strategies provide a continuum of object-level results that grant a system designer or a real-time metareasoner wide ranges over which to optimize the duration of reasoning before acting. The flexible strategies also allow us to squeeze the most inference out of a problem in situations where

---

*This work was supported by a NASA Fellowship under Grant NCC-220-51, by the National Science Foundation under Grant IRI-8703710, by the National Library of Medicine under Grant R01LM0429, and by the U.S. Army Research Office Grant P-25514-EL. Computing facilities were provided by the SUMEX-AIM Resource under NIH Grant RR-00785.

there is uncertainty in a deadline by allowing us to continue to refine a result until a deadline arrives.[1]

We have pursued the development of flexible decision-theoretic inference by decomposing difficult problems into sets of subproblems, and configuring representations and inference strategies that allow the most relevant portions of the problem to be analyzed first. We have worked to develop calculi for managing the error or uncertainties about the answer, in light of the current state of incompleteness of the analysis. Such techniques apply knowledge about the logic of the solution method and partial characterization of the problem instance at hand to generate answers with bounds or probability distributions over the answers that would be calculated if sufficient resources were available.

We can reformulate a complex problem into a set of smaller, related problems, by modulating the completeness of an analysis or the level of abstraction of the distinctions manipulated by the analysis. Previous work on the modulation of abstraction has examined the reformulation of an optimal set of distinctions into a hierarchy of abstractions at different levels of detail [11]. These techniques have been used for simplifying value-of-information analyses for generating recommendations in the Pathfinder system for assistance with pathology diagnoses [6]. In this paper, we explore the modulation of completeness of probabilistic inference by decomposing a problem into a set of inference subproblems, and by ordering the solution of these problem components by their expected relevance to the final belief. Each subproblem represents a plausible context. This work can be viewed as a reformulation of an all-or-nothing approach to probabilistic inference in belief networks. As opposed to generating exact probabilities about a proposition of interest, we seek to generate distributions or logical bounds on probabilities by keeping track of the contexts that we have not considered. Our method, called *bounded conditioning*, is a graceful analog to the method of conditioning for probabilistic entailment in a belief network, developed by Pearl [13].

## 2 Complex Probabilistic Inference

We have been investigating decision-theoretic inference under bounded resources within the Protos project. We represent decision problems with the belief-network representation. In a belief network, nodes represent propositions of interest, and arcs represent dependencies among belief in the nodes. In our medical decision systems, we often must deal with large, multiply connected networks. We know that the complexity of probabilistic inference within belief networks is $\mathcal{NP}$-hard [4]. Figure 1 pictures a belief network that represents distinctions and probabilistic relationships in the intensive-care unit (ICU)[2]. The ICU network is multiply connected and contains 37 nodes.

### 2.1 The Method of Conditioning

There is a variety of approximate and exact methods for performing inference with the belief-network representation. A recent survey of alternative methods is found in [9]. We will review the *method of conditioning* [13].

In the method of conditioning, dependency loops in a belief network are broken by a set of nodes called a *loop cutset*, so named because its members are selected such that every loop (a minimal multiply connected subset of the network) is cut by at least one member of the set. After the loop cutset is identified, the method of conditioning requires the instantiation of the members of the cutset. Combinations of instantiations of the loop-cutset nodes are *instances* of the cutset.

---

[1] See [8] for a discussion of flexible reasoning and decision-theoretic optimization in the context of basic computational tasks such as sorting a file of records.

[2] The prototype network, called ALARM, was designed and assessed by Ingo Beinlich [1]

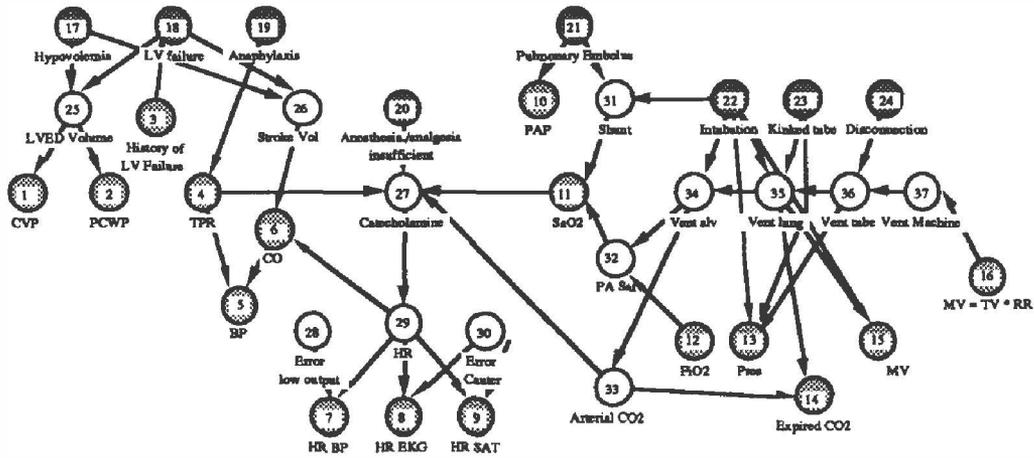

Figure 1: A multiply connected belief network representing the uncertain relationships among some relevant propositions in the intensive-care unit (ICU).

In the context of some observed evidence, the instances are solved with an efficient method for solving singly connected networks. We apply a distributed algorithm for solving singly connected networks, developed by Kim and Pearl [12]. For singly connected belief networks, this algorithm is linear in the size of the network. Finally, in the method of conditioning, the answers of the singly connected subproblems are combined to calculate a final probability of interest.

The number of instances is equal to the product of the number of values of each node in the loop cutset. That is, the number of instance subproblems that must be solved is $\prod_{i=1}^{m} V(C_i)$ where $C_i$ is a node in the cutset, and $V(C_i)$ is the number of possible outcomes for $C_i$; for binary-valued variables, the number of instances that must be solved is $2^{|\text{cutset}|}$. As this number grows exponentially with the size of the cutset, it is important to identify the smallest possible cutset. Problems, approximations, and empirical testing of means of identifying good cutsets in complex networks are discussed in [15].

When evidence is observed, the method of conditioning calls for the propagation of this evidence in each instance in order to calculate the updated posterior probabilities for the nodes of network. We associate with each unique instance $c_1 \ldots c_m$ an integer label $i$, and designate $p(c_1 \ldots c_m)$ as the weight of the instance $w_i$. The weights for all instances are calculated and stored during initialization of the priors in the network. Initially, therefore,

$$w_i = p(c_1 \ldots c_m) = p(c_1)p(c_2|c_1)\ldots p(c_m|c_1 \ldots c_{m-1})$$

where the loop cutset consists of $n$ nodes, $C_1 \ldots C_m$, and these have been assigned values $c_1 \ldots c_m$ in instance $i$.

During computation for initializing the network, we calculate, for each cutset instance, the marginal probabilities for each node in the network, given the values assigned to the loop-cutset nodes in that instance. Therefore, for each value $x$ of node $X$, and for each instance $i$, we have

$$p(x|\text{instance } i) = p(x|c_1 \ldots c_m)$$

Thus, calculating $p(x)$,

$$p(x) = \sum_{c_1,\ldots c_m} p(x|c_1 \ldots c_m)p(c_1 \ldots c_m) = \sum_i p(x|\text{instance } i) \times w_i$$

where $i$ is iterated through all $n$ instances of $c_1 \ldots c_m$.

When we discover the truth status of one or more propositions in the network, as is the case when we observe evidence, we first update the weights for all loop-cutset instances such that, together, they still are equal to the joint probability of the loop-cutset nodes and the evidence. We do this update for each instance subproblem by multiplying the current weight of each instance by the probability of the evidence given that instance. If we observe value $e$ of node $E$, then we calculate the new weight, $w_i^*$, of instance $i$ as follows:

$$w_i^* = p(c_1 \ldots c_m | e) = \alpha p(c_1 \ldots c_m, e) = \alpha p(e | c_1 \ldots c_m) p(c_1 \ldots c_m)$$

$$= \alpha p(e | \text{instance } i) \times w_i$$

where $\alpha = \frac{1}{p(e)}$, obtained by normalizing the new weights.

After the new weights are calculated, we apply Pearl's algorithm for propagating evidence in a singly connected network [13] to solve each instance. For each instance $i$, we assign a probability to each value $x$ of node $X$,

$$p(x | e, \text{instance } i) = p(x | e, c_1 \ldots c_m)$$

This, in turn, allows us, at any time, to obtain $p(x|e)$ for any node; we simply sum the belief over all instances, weighted by the likelihood of the instances:

$$p(x|e) = \sum_{c_1 \ldots c_m} p(x|e, c_1 \ldots c_m) p(c_1 \ldots c_m | e) = \sum_i p(x|e, \text{instance } i) \times w_i^*$$

For additional evidence, we repeat this procedure, each time multiplying the old weight assigned to an instance by the probability of the observed value given that instance. Thus, the method of conditioning provides a mechanism for performing general probabilistic inference in multiply connected belief networks.

## 3  Bounded Conditioning

Recall that the computational complexity of the method of conditioning is an exponential function of the size the cutset. If we have a large number of cutset instances, exact probabilistic inference using this method may not be feasible in situations where sufficient time is not available or where delay is costly. To provide computation that has maximal value to a computational decision system or system user, we must consider the net benefits of computation in the context of the costs of reasoning. See [8] for a discussion of prototypical contexts of resource cost and availability.

### 3.1  Directing Inference by Relevance

With *bounded conditioning*, rather than being constrained to wait until a point probability is generated, we seek to determine, with less computation, that the "probability of interest"—the probability that would be calculated with infinite computation—is above or below a certain value. We consider each loop-cutset instance as a separate subproblem, representing a plausible context, and generate bounds on probabilities of interest by accounting for the contexts that have not yet been explored. There have been several previous research efforts on algorithms that generate bounds on probabilities [3,14]. With bounded conditioning, we obtain exact upper and lower bounds on the probability of each value of each node in the network based on characterizing the maximal positive and negative contributions of the unsolved instances. We continually probe the unexplored portion of the reasoning problem to order the analysis of instance subproblems by their expected contribution on the tightening of bounds.

The bounded-conditioning approach was designed originally for exploring principles of inference under varying limitations in reasoning resources. Probability bounds and information about the expected convergence of bounds can be used to tell us about optimal deliberation before action, given the costs of computation. The application of flexible problem solving in systems that perform metareasoning about the value of continuing to perform inference, versus acting in the world, is discussed in [10].

## 3.2 A Partial-Instance Bounding Calculus

With bounded conditioning, after observing some evidence, we (1) calculate the weights assigned to each subproblem, (2) sort the subproblems by weight, (3) update each instance in sequence, and (4) integrate the results of each loop-cutset instance, based on the weights of the instances and knowledge about the unexplored portion of the problem.

During the offline initialization of the network, we calculate the marginal probabilities over the values of nodes in the network and the prior weight $w_i$ on each instance $i$. After initialization, and after bounded conditioning has been used to solve completely an inference problem (e.g., the calculation of a point probability), the network is in a completely solved state. We will address updating the complete states first. Later, we will generalize the bounding calculus to handle new evidence in the context of incompletely solved networks.

**Inference from a Complete State.** Let us first consider the case, where a fully-initialized belief network is updated, given the observation of a piece of evidence. After observing value $e$ of evidence node $E$, we first recalculate the new loop-cutset weights $w_i^*$ for each instance $i$ in the context of the evidence. Next, we solve the marginal probabilities of values of nodes in each instance, in order of the prior weight of the instances.

For those loop-cutset instances that we have updated, we know $p(x|e, \text{instance } i)$ with certainty. For the loop-cutset instances that we have not yet updated, we know with certainty that $0 \leq p(x|e, \text{instance } j) \leq 1$. Therefore, for any node $X$ and value $x$, we can obtain a lower bound on $p(x|e)$ by substituting 0 for those probabilities we have not yet calculated. We can calculate an upper bound on $p(x|e)$ by substituting 1 for these probabilities, as this is the maximal contribution of the probability of seeing the evidence given an instance.

Let us assume that we only propagate the evidence through the network for a subset of instances 1 through $j$; therefore, we do not update the probabilities for instances $j+1$ through $n$. After propagating the evidence for instances 1 through $j$, we can calculate bounds on $p(x|e)$ as follows:

$$\begin{aligned}
\text{Lower bound on } p(x|e) &= \sum_{i=1}^{j} p(x|e, \text{instance } i) \times w_i^* + \sum_{i=j+1}^{n} 0 \times w_i^* \\
&= \sum_{i=1}^{j} p(x|e, \text{instance } i) \times w_i^* \quad (1)
\end{aligned}$$

Similarly, for the upper bound,

$$\begin{aligned}
\text{Upper bound on } p(x|e) &= \sum_{i=1}^{j} p(x|e, \text{instance } i) \times w_i^* + \sum_{i=j+1}^{n} 1 \times w_i^* \\
&= \sum_{i=1}^{j} p(x|e, \text{instance } i) \times w_i^* + \sum_{i=j+1}^{n} w_i^* \quad (2)
\end{aligned}$$

Thus, the difference between these bounds is

$$\text{Upper bound} - \text{Lower bound} = \sum_{i=j+1}^{n} w_i^*$$

Note that the size of the bounds interval is equal for the posterior probabilities for the values of all nodes in the network, and depends only on the weight of the unexplored problem. The case of performing bounded conditioning from a complete state is appropriate in situations where evidence is seen at intervals long enough to allow complete updating, yet where decisions may have to be made as soon as possible after the observation of that evidence.

**Bounding from an Incomplete State.** We now generalize our bounding calculus to allow us to update a network with new evidence before previous evidence has been completely analyzed. Recall that the revised weight for an instance, in light of new evidence, is obtained by multiplying the old weight for that instance by the probability of the observed evidence in that instance, and normalizing the product by dividing the result by the marginal probability of the evidence. To compute the weights, we must first calculate the marginal probabilities within each instance. If we did not update the belief in values of the nodes in a particular instance when we added the last piece of evidence, it is not possible to obtain the probability of the new evidence, since these instances have not been updated. To reason about the relevance of additional pieces of evidence, given a previously incomplete analysis of a subset of instance subproblems requires us to apply a bounding analysis to the weights themselves. This makes our bounding calculus a bit more complicated.

We are seeking $p(\text{instance } i|e, f)$ for all instances. Suppose we previously considered evidence $e$ for node $E$, and we solved instances 1 through $j$ out of a total of $n$ instances. Now, we observe value $f$ for node $F$. To calculate the new instance weights given evidence $f$, we can no longer simply normalize $p(f, \text{instance } i|e)$ over all instances $i$ because we have not yet updated all of the instances. However, we can calculate bounds on these weights. For instances 1 through $j$, the belief in the conjunction of the new evidence and the old is calculated as follows:

$$p(f, \text{instance } i|e) = p(f|e, \text{instance } i) \times p(\text{instance } i|e)$$

If we knew this for all instances $i$, we would normalize to obtain the instance weights; that is

$$w_i^* = p(\text{instance } i|e, f) = \frac{p(f, \text{instance } i|e)}{\sum_k p(f, \text{instance } k|e)}$$

Since we do not know $p(f, \text{instance } k|e)$ for all instances $k$, we can assign bounds to $w_k^*$ by taking into account that for all $k$,

$$0 \leq p(f, \text{instance } k|e) \leq p(\text{instance } k|e) \tag{3}$$

We can generate lower and upper bounds on the weights by normalizing with factors that we know are larger and smaller, respectively, than $\sum_{i=1}^{n} p(f, \text{instance } i|e)$. For instances 1 through $j$,

$$\frac{p(f, \text{instance } i|e)}{\sum_{k=1}^{j} p(f, \text{instance } k|e) + \sum_{k=j+1}^{n} p(\text{instance } k|e)} \leq w_i^* \leq \frac{p(f, \text{instance } i|e)}{\sum_{k=1}^{j} p(f, \text{instance } k|e)} \tag{4}$$

Let us now turn to the calculation of the revised weights $w_i^*$ of instances $j + 1$ through $n$. We need only to compute an upper bound for the weights of these instances for use in determining the upper bounds on posterior probabilities. Let $w_k^L$ indicate a lower bound on the weight of instance $k$ in instances 1 through $j$ (left side of Equation 4), $w_k^U$ indicate an upper bound on the weight in instances 1 through $j$ (right side of Equation 4), and $w_k^{U'}$ indicate an upper bound on the weight of instances $j+1$ to $n$. Since $\sum_i w_i^* = 1$, we know that

$$\left[1 - \sum_{k=1}^{j} w_k^U\right] \leq \sum_{i=j+1}^{n} w_i^* \leq \left[1 - \sum_{k=1}^{j} w_k^L\right] \tag{5}$$

Equation 3 justifies calculation of upper bounds on revised instance weights through normalizing the previous weights $w_i$ of instances $j+1$ through $n$. Noting that $\sum_{k=1}^{j} w_k^L \leq \sum_{k=1}^{j} w_k^*$ and $1 - \sum_{k=1}^{j} w_k^U \leq \sum_{k=j+1}^{n} w_k^*$ (from Equation 5), we can obtain upper bounds $w_i^{U'}$ on the revised weights $w_i^*$ of each instance $j+1$ through $n$ as follows:

$$w_i^{U'} = \frac{w_i}{\sum_{k=1}^{j} w_k^L + \left[1 - \sum_{k=1}^{j} w_k^U\right]} \tag{6}$$

Let us assume that for the new evidence $f$, we update only instances 1 through $h$ where $h \leq j$. As instances $j+1$ through $n$ were not updated when the last piece of evidence was added, they cannot be updated now. Thus, we update and sort only instances 1 through $h$. Bounds on the posterior probability of interest are obtained in a similar manner as in Equations 1 and 2. Let $P^L(x|e, f)$ and $P^U(x|e, f)$ represent, respectively, the lower and upper bounds on the posterior probability. After updating instances 1 through $h$ with the new evidence, we have the following bounds on $p(x|e, f)$:

$$\begin{aligned}
P^L(x|e, f) &= \sum_{i=1}^{h} p(x|e, f, \text{instance } i) \times w_i^L \\
P^U(x|e, f) &= \sum_{i=1}^{h} p(x|e, f, \text{instance } i) \times w_i^U \\
&\quad + \sum_{i=h+1}^{j} w_i^U + \sum_{i=j+1}^{n} w_i^{U'}
\end{aligned}$$

## 4  Convergence: Two Theoretical Scenarios

We can gain insight into the behavior of bounded conditioning by making assumptions about the distribution of belief over instances. We examine the case of updating a probability given complete prior initialization of the instance weights. We discuss rates of convergence of probabilistic reasoning on two prototypical distributions over weights, $w_i$, assigned to instance subproblems.

**Worst-Case Convergence.** Consider a belief network that is cut by a loop cutset of $n$ binary nodes. The degree of symmetry versus asymmetry in the way mass is apportioned over the values of nodes in the cutset of a belief network is relevant to the convergence behavior of bounded conditioning. In the worst case for the convergence of bounded conditioning in the context of available weights, all subproblem instances have the same weight. With $n$ binary nodes, we have $2^n$ instances, each with weight $2^{-n}$. According to our convergence calculus, described in Section 3, at time $t$, our bounds will be described by $1 - (2^{-n} \times \frac{t}{k})$ where $k$ is the amount of time required by each instance for solution. Figure 2 shows an example of worst-case linear convergence for a loop cutset of 15 nodes. This linear convergence is the slowest rate of refinement we can expect with bounded conditioning. Even in such worst cases, the utility structure of the decision problem—for which the inference is being performed—can dictate that we need to solve only a portion of the entire inference problem to derive a great fraction of the value of perfect inference [7,10].

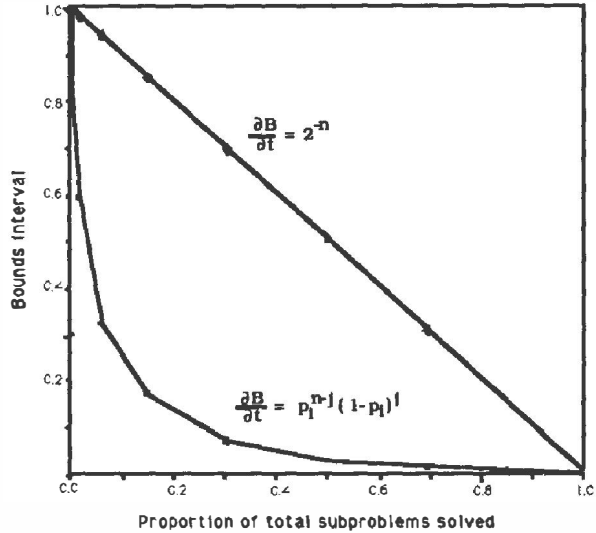

Figure 2: The linear, worst-case (upper curve) and better-case (lower curve) convergence of bounds for a marginally independent loop cutset consisting of 15 nodes. The better-case convergence is based on an assumption of homogeneous asymmetry in the distribution over the probability of values for each node $(0.75, 0.25)$.

**Better Performance.** Asymmetric distributions over the conditional probabilities of alternative values of specific cutset nodes, given values assumed for other nodes in a cutset, allow for a wide range of differences among the weights for instances. Sorting and sequentially solving these subproblems enables a reasoner to take advantage of the nonlinearity in weights with subproblems. Such situations often enable a reasoner to capitalize on a disproportionate amount convergence for early computation.

Consider the case where we again have a loop cutset of $n$ binary nodes. Now, however, we have an identical asymmetric contribution for values of each node in the cutset, within each instance. Each node takes on the value true with probability $p$, and the value false with probability $1 - p$. Within such a network, we have several sets of instances with equal weight. We are assuming that the dependence among cutset nodes is insignificant to this analysis. In particular, we have sets of instances with weight

$$p^{n-j}(1-p)^j$$

each of cardinality $\frac{n!}{j!(n-j)!}$, from the largest to smallest weights as $j$ varies from 0 to $n$. The bounds interval, based on an incomplete analysis, in this case is described by

$$1 - \sum_{j=0}^{m \leq n} \frac{n!}{j!(n-j)!} p^{n-j}(1-p)^j$$

where $m$ describes the last set of evidence of a particular weight to be evaluated.

In juxtaposition to a scenario of worst-case convergence, Figure 2 displays a better-case convergence, where each of the 15 loop-cutset nodes takes on the value true with probability 0.75, and the value false with probability 0.25. This graph shows a piecewise-linear convergence at different rates for each value of $j$. The rate of convergence with the solution of subproblems is maximal at the outset of inference. The proportions of total problem instances analyzed ($2^{15} = 32,768$) are listed on the $x$ axis. Although we have not included the possible effects of dependencies among cutset nodes that may exist in a real belief network, the independence analysis for the weight distributions for the case of the homogeneous loop cutset can give us intuition about the convergence of bounded conditioning on distributions over instance weights from real networks.

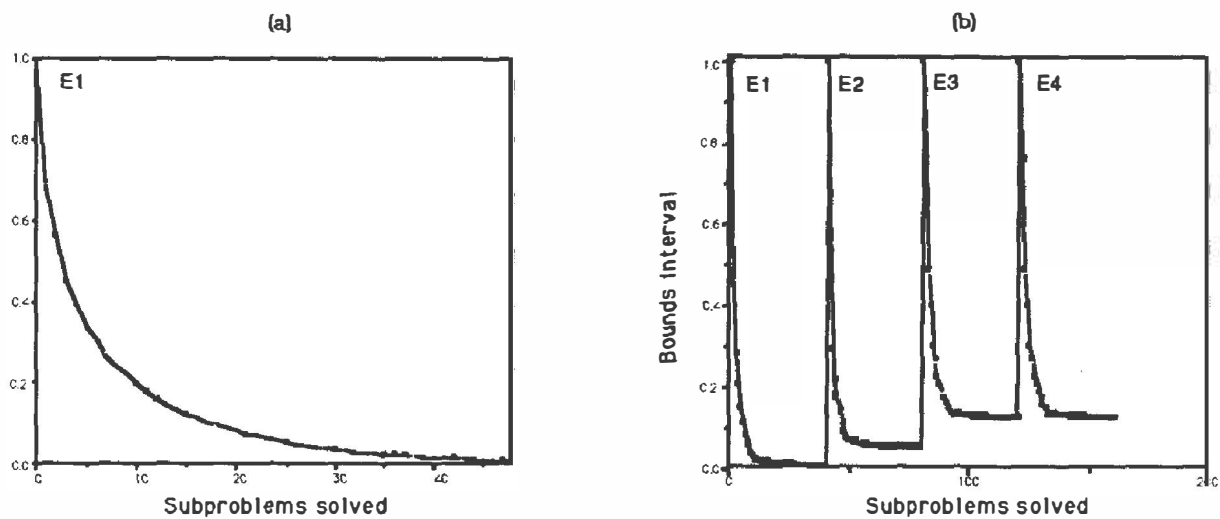

Figure 3: (a) The empirical convergence of the width of bounds on a probability of interest in response to a single observation with the application of bounded conditioning to the ICU network pictured in Figure 1. Each instance subproblem requires approximately 5 seconds of computation. (b) The convergence of the bounds interval in response to a sequence of 4 observations.

## 5 Empirical Performance

We have performed preliminary analysis of bounded conditioning on experimental belief networks constructed with a belief-network generator and on the ICU network described in Section 2. We studied several loop cutsets for the ICU network. A sample loop cutset consists of 5 nodes that leads to 108 different singly connected network subproblems. Figure 3(a) shows the typical form of the convergence displayed by bounded conditioning when updating all nodes in the ICU network, given an observation. For this network, the solution of each instance subproblems require approximately 5 seconds of computation. The graph shows the bounds on the probability of a proposition of interest with the solution of additional subproblems. We found that the bounds for a large set of updates for this problem decay at rate that can be modeled approximately with a negative exponential, $e^{-k(t+1)}$, with different decay constants $k$. This convergence information has been used in reasoning about the expected value of continuing to apply the bounding algorithm in the context of the costs and benefits of acting given different states of the world [10].

Figure 3(b) shows an example of the application of the more complicated bounding scheme for handling the general case of new evidence impinging on a reasoner while previous bounded conditioning is in progress. In this example, we consider the updating of 4 observations, with each of the new observations becoming available after the Protos reasoner completes 40 out of the total 108 subproblems (about 35% of each update task) required to completely solve each evidential update. Notice that in both cases, a great portion of the inference problem is solved for a relatively small fraction of the work required for a complete analysis.

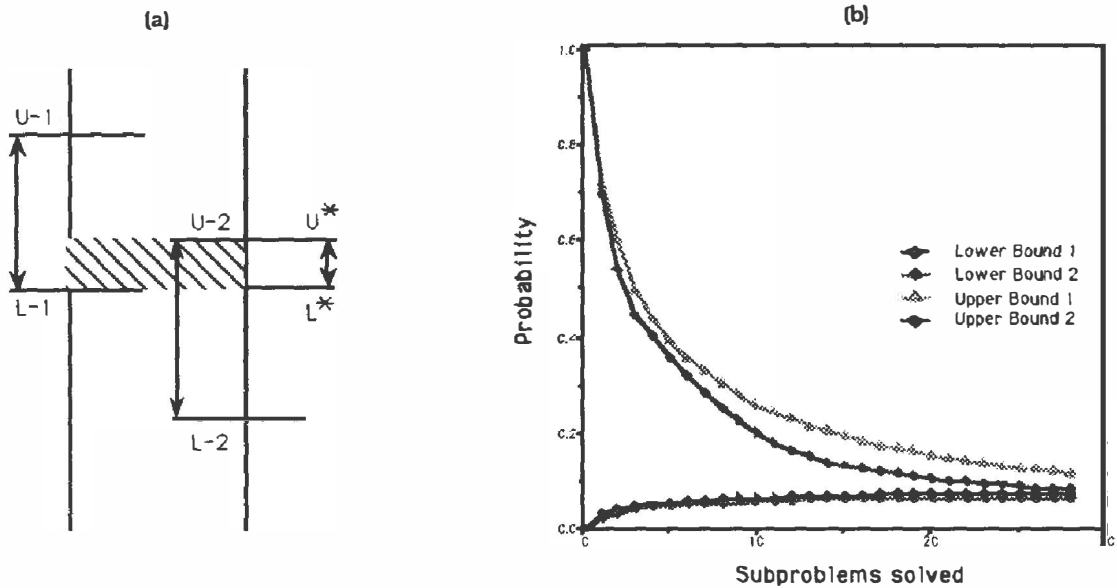

Figure 4: (a) The goal of concurrent bounded conditioning is to solve simultaneously different bounding problems, and to combine the different bounds (U-1, L-1 and U-2, L-2) to construct final bounds ($U^*$, $L^*$) on a probability of interest. (b) Upper and lower bounds associated with two concurrent bounded-conditioning analyses of an ICU inference problem

## 6 Research on Bounded-Conditioning

We are continuing to explore the empirical behavior of bounded conditioning for complex belief network problems, with a particular focus on the ability of the algorithm to deliver valuable computation in the context of the utility structure of decision models that make use of computed beliefs. We are also examining extensions of the fundamental algorithm.

**Concurrent Bounded Conditioning.** There are typically several loop cutsets for a belief network. Alternative cutsets update the upper and lower bounds of a probability of interest in a different manner. In preliminary experimentation, we investigated gains that could be ascertained through the concurrent processing of a belief network with two different bounded-conditioning analyses. As indicated in Figure 4(a), given a set of two or more bounding analyses, we select the greatest lower bound and the lowest upper bound to construct final upper and lower bounds on a probability of interest. The less the overlap among different sets of bounds, the greater the benefits of concurrency. Figure 4(b) displays graphs of analyses based on two different loop-cutset. The convergence of the final bounds is just the interval between the greatest lower bound and the smallest upper bound. In analyses to date, the overhead of executing two different problems is greater than the gains. In many other cases, however, this may not be true. We plan to investigate this issue further. We have also completed a promising parallel implementation of bounded conditioning and will continue to test this approach.[3]

---

[3]The parallel implementation was developed by Adam Galper, a doctoral student in our laboratory.

**Control of Bounded Conditioning.** For the case of generalized bounding from an incomplete state, there is opportunity for making problem-solving control decisions about the degree to which an earlier evidential subproblem should be analyzed before considering new observations. We are investigating this tradeoffs at the microstructure of bounded conditioning. An area of interest is the use of compiled control rules based on the network structure, and on previous and expected new observations. We see opportunity for applying a real-time greedy search along alternative dimensions of updating, to optimize the rate of bounding.

**Idle-Time Computation.** We wish to make the best use of the idle-time between completion of a previous update and the observation of new evidence. In many applications, there may be time between observations for planning a best approach to future updates. For example, our reasoner can begin to bound as yet unobserved evidence expected in the future with high probability. This approach would ready a reasoner for future reflection and action.

**Compilation.** There is opportunity to compile and make available such useful information as the instance weights and the probability of evidence, conditioned on previous sets of observed evidence. We are exploring strategies for precomputing and caching these probabilities based on the importance and commonality of the evidence.

**Simulation for Calculating Instance Weights.** Finally, there can be great synergy in the application of alternative algorithms for performing inference about components of the bounded conditioning problem. We are examining the integration of stochastic simulation methods [2] for enumerating the cutset weights in light of new evidence. We believe that a more robust version of bounded conditioning will result from this work.

## 7 Summary

We introduced the bounded conditioning method for probabilistic inference and described how the strategy can be used to generate bounds on the marginal probabilities for any node in a belief network. Bounded conditioning exhibits the useful properties of continuity, monotonicity, and convergence, enabling a reasoner to exchange arbitrary quantities of computational resource for incremental convergence on probability bounds. The approach, and its future extensions, promise to be useful in reasoning under the general conditions of uncertainty in available reasoning resources. We have found, in complex belief networks, that the method solves a great portion of a probabilistic inference problem with a solution of a fraction of the total number of instance subproblems. Future efforts will seek to characterize the method more fully and to take advantage of the structure of particular belief networks. Areas of opportunity for additional research include concurrent bounded conditioning with multiple cutsets, control strategies for optimizing the convergence of bounds, strategies for guiding idle-time reasoning about expected updating, the compilation and caching of instance weights, and the use of complementary approximation methods for updating the weights on instance subproblems.